\documentclass[10pt,twocolumn,letterpaper]{article}

\usepackage{cvpr}
\usepackage{times}
\usepackage{epsfig}
\usepackage{graphicx}
\usepackage{amsmath}
\usepackage{amssymb}
\usepackage{subcaption}

\usepackage[pagebackref=true,breaklinks=true,letterpaper=true,colorlinks,bookmarks=false]{hyperref}

\cvprfinalcopy 

\newcommand{\gcj}[1]{{{\color{darkred}(Chenjie: #1)}}}
\newcommand{\ignore}[1]{}
\newcommand{\tbd}[1]{#1}

\renewcommand{\tbd}{\ignore}
\newcommand{\libvpx}{{\fontfamily{cmss}\selectfont libvpx} }
\newcommand{\libvpxx}{{\fontfamily{cmss}\selectfont libvpx}}

\newcommand{\myfigref}[1]{Figure~\ref{f:#1}}
\newcommand{\eqnref}[1]{Equation~(\ref{eqn:#1})}

\newcommand\blfootnote[1]{%
  \begingroup
  \renewcommand\thefootnote{}\footnote{#1}%
  \addtocounter{footnote}{-1}%
  \endgroup
}

\begin{document}
\title{Neural Rate Control for Video Encoding using Imitation Learning}

\author{Hongzi Mao\textsuperscript{\dag}\textsuperscript{$\star$}\;\; Chenjie Gu\textsuperscript{\dag}\;\; Miaosen Wang\textsuperscript{\dag}\;\; Angie Chen\textsuperscript{\dag}\;\; Nevena Lazic\textsuperscript{\dag}\;\; Nir Levine\textsuperscript{\dag}\\ Derek Pang\textsuperscript{*}\;\; Rene Claus\textsuperscript{*}\;\; Marisabel Hechtman\textsuperscript{*}\;\; Ching-Han Chiang\textsuperscript{*}\;\; Cheng Chen\textsuperscript{*}\;\; Jingning Han\textsuperscript{*}\;\; \\
\\
\textsuperscript{\dag}DeepMind \;\;\;\;\;\; \textsuperscript{*}Google \;\;\;\;\;\; \textsuperscript{$\star$}MIT CSAIL\\
}

\maketitle

\begin{abstract}
In modern video encoders, rate control is a critical component and has been heavily engineered.
It decides how many bits to spend to encode each frame, in order to optimize the rate-distortion trade-off over all video frames. 
This is a challenging constrained planning problem because of the complex dependency among decisions for different video frames and the bitrate constraint defined at the end of the episode.

We formulate the rate control problem as a Partially Observable Markov Decision Process (POMDP), and apply imitation learning to learn a neural rate control policy. We demonstrate that by learning from optimal video encoding trajectories obtained through evolution strategies, our learned policy achieves better encoding efficiency and has minimal constraint violation. In addition to imitating the optimal actions, we find that additional auxiliary losses, data augmentation/refinement and inference-time policy improvements are critical for learning a good rate control policy.  We evaluate the learned policy against the rate control policy in \libvpxx, a widely adopted open source VP9 codec library, in the two-pass variable bitrate (VBR) mode.
We show that over a diverse set of real-world videos, our learned policy achieves 8.5\% median bitrate reduction without sacrificing video quality.

\end{abstract}
\section{Introduction}
\label{s:intro}
\blfootnote{\textsuperscript{$\star$}Work done during the internship at DeepMind. Correspondence to Chenjie Gu  \textless\url{gcj@google.com}\textgreater.}
%
In recent years, video traffic keeps growing and constitutes over 75\% of the Internet traffic~\cite{cisco-internet-report}.
Efficient video encoding is key to improving video quality and alleviating backbone Internet traffic pressure for video tasks such as video on demand (VOD) and live streaming decoding.

In modern video encoders, \emph{rate control}
is a critical component that directly modulates the trade-off between rate (video size) and distortion (video quality).
The rate control module manages the distribution of available bandwidth, usually constrained by the network condition. It directly determines how many bits to spend to encode each video frame, by assigning a quantization parameter (QP). The goal of rate control is to maximize the encoding efficiency (often measured by the Bjontegaard delta rate) as well as maintaining the bitrate under a user-specified target bitrate.

Rate control is a constrained planning problem and can be formulated as a Partially Observable Markov Decision Process. For a particular frame, the QP assignment decision depends on (1) the frame's spatial and temporal complexity, (2) the previously encoded frames this frame refers to, and (3) the complexity of future frames. In addition, the bitrate constraint imposes an episodic constraint on the planning problem. 
The inter-dependency of encoding decisions and the episodic constraint make it challenging to engineer optimal rate control algorithms.

To evaluate the potential of improving the rate control policy, we conducted an experiment to measure how much we may increase the Bjontegaard delta rate (BD-rate) compared to the rate control policy in \libvpxx, a widely adopted open source VP9 codec for adaptive video streaming services, such as YouTube and Netflix. We collected 1600 videos spanning diverse content types and spatial/temporal complexities. For every video, we applied an evolution strategy algorithm which maximizes the peak signal-to-noise ratio (PSNR) under \libvpxx's two-pass variable-bitrate (VBR) mode.
We observe an average 13\% improvement in BD-rate, indicating a big gap between the \libvpxx's rate control policy and an optimal rate control policy.

In this paper, we use imitation learning to train a neural rate control policy and demonstrate its effectiveness in improving encoding efficiency.
We first generate a teacher dataset by searching for optimal QP sequences for individual videos. We then train a neural network by imitating these optimal QP sequences. However, imitating the optimal QP sequences by itself fails to learn a good rate control policy.
In order for the policy to generalize to many videos and to satisfy the bitrate constraint, we design additional auxiliary losses and customized hindsight experience replay to train the policy. During inference, we use a truncation trick, a simple technique to prune spurious action values and improve encoding efficiency. Finally, we augment the learned policy with feedback control, in order to precisely hit the target bitrate, and therefore minimize constraint violations.

Although the method is developed with the goal of solving the rate control problem, it is general and is applicable to other constrained planning problems and imitation learning methods. For example, the feedback control augmentation defines a policy improvement operator for a class of constrained planning problems, and may be used in reinforcement learning.

We evaluate and compare our learned neural rate control policy against the default rate control policy in \libvpxx's two-pass VBR mode. When evaluated on 200 real-world videos never seen during training, our learned policy achieves a 8.5\% median reduction in bitrate while maintaining the same video quality.

In summary, our contributions include:
\begin{enumerate}
    \item We formulate the rate control problem as a Partially Observable Markov Decision Process (POMDP) (\S\ref{s:problem});
    \item We develop an imitation learning method, including teacher dataset generation, auxiliary losses, model architecture and a customized hindsight experience replay method, for learning a rate control policy (\S\ref{s:design});
    \item We develop a policy improvement operator that augments a learned policy with feedback control, and show that it significantly reduces the constraint violation (\S\ref{s:design-control});
    \item We show that our learned rate control policy achieves 8.5\% median reduction in bitrate while maintaining the same video quality (measured in PSNR) (\S\ref{s:eval}).
\end{enumerate}

    \tbd{
    \item An open sourced dataset of our imitation target QP sequences for future research (Appendix~\ref{a:open_source}). \gcj{tbd}
    }

\tbd{

\section{Introduction (V2, Hongzi)}
%
Recent years have witnessed a clear domination of video traffic on the entire Internet, reaching over 75$\%$ of the total traffic volume~\cite{cisco-internet-report}. Video encoding sits on the forefront of the video delivery pipeline. Efficient video encoding is the key to improving video quality and alleviating backbone Internet traffic pressure for all downstream video tasks such as decoding and streaming.
%
In this paper, we focus on the \emph{rate control} problem at the core of the video encoding process~\cite{rate-control-overview, rate-control-patent}. At a high level, rate control decides how many bits to assign to each video frame through a quantization parameter (QP). Assigning more bits to a frame generally increases the frame quality. The goal is to optimize a constrained objective: achieving the highest overall frame quality of a video while ensuring the total bits spend is within a budget limit.

In essence, this rate control task is a bit budget allocation problem for each video chunk (see \S\ref{s:problem} for details). The fundamental challenge of this problem is that the mapping from the bit assignment to the video quality depends on (1) the visual content of a frame (e.g., over-assigning bits to a largely empty frame would not improve the quality much), (2) the complex relation among consecutive video frames (e.g., encoding a key frame well in a static scene also improves the quality of subsequent frames) and (3) the planning over the entire video chunk (e.g., proactively preserving bits for lateral frames if current simple frames only need minimal bits to encode). Facing these challenges, existing approaches resort to fixed  heuristics that do not tailor for each individual video content and they leave significant room for performance improvement.

To quantify the room for improvement, we leverage the parallel compute power to optimize the QP sequence for each video \emph{individually and independently} using Evolution Strategies (ES)~\cite{es}. We treat each QP action as a ``parameter'' in the ES search; essentially, we offline search for the optimal QPs in a brute force manner (Appendix \ref{a:es}). We found that ES outperforms the state-of-the-art VP9 encoder in terms of ``projected video quality''---\,improving the video quality while using the same amount of bits. 

In practice, however, ES alone is not scalable to find optimal QPs for each video in production. We therefore formulate the rate control problem in the Markov decision process (MDP) problem structure and use imitation learning to mimic the optimal QP selection policy.
Off-the-shelf imitation learning is not enough to train a strong QP selection policy that generalizes to unseen videos. Once deviates from the target, vanilla imitation cannot navigate back to the ``normal'' trajectory (i.e., following QP actions from the target policy)~\cite{imitation-survey}. In our design, we propose a series of techniques to resolve this issue for the video encoding application (\S\ref{s:design}). Specifically, we employ an auxiliary loss to regularize the bits spend to hit the target. Also, we use a customized hindsight experience replay~\cite{her} technique to regenerate the QP sequences that are easier to imitate especially in the case of trajectory deviation. Moreover, at inference time, we use truncation trick and propose a model-based feedback control method to guardrail the bit consumption trajectory. Importantly, this method modifies the QP action \emph{based on the learned model} and it empirically preserves the encoding performance while making sure that the bit consumption hits the target.

We integrated our learning-based rate controller with the existing VP9 codec and tested on real world videos from one of the largest video platform (\S\ref{s:eval}). Over 200 testing videos in production, spanning across gaming, films and natural scenery video genres, our approach achieves $10.6\%$ reduction in bitrate while achieving the same quality. Also, we performed a series of ablation studies to understand the impact of each proposed technique. Importantly, this problem also paves an relatively easy path to video encoding in production systems: by only changing the QP assignment logic, this modification is \emph{not} as intrusive as other end-to-end learning systems that swap out the entire control pipeline~\cite{grl1, grl2, placeto, rl-cc-icml, pensieve, decima}.
%
To promote more research in the direction of learning-based rate control and other related tasks in video encoding, we open source the ES target QP sequences, as well as the intermediate searched QP sequences, for videos in the public YouTube UGC dataset~\cite{ugc}.

In summary, the contributions of this paper include:
\begin{CompactEnumerate}
    \item An MDP framework of the rate control problem, including the frame features as state observations and the reward formulation for quality improvement and bit budget control (\S\ref{s:problem});
    \item A series of techniques to generalize the imitation learned rate control policy to unseen videos. This includes a regularizer design, a hindsight experience replay method to regenerate easier-to-train data, and a feedback control mechanism to guardrail the bits spend in inference (\S\ref{s:design});
    \item Integrating the learning-based rate controller in production video encoder to achieve $10.6\%$ reduction in bit consumption while providing the same video quality (\S\ref{s:eval});
    \item An open sourced dataset of our imitation target QP sequences for future research (Appendix~\ref{a:open_source}).
\end{CompactEnumerate}

}

\begin{figure*}[tbp]
\centering
\begin{subfigure}{0.60\textwidth}
\includegraphics[width=\textwidth]{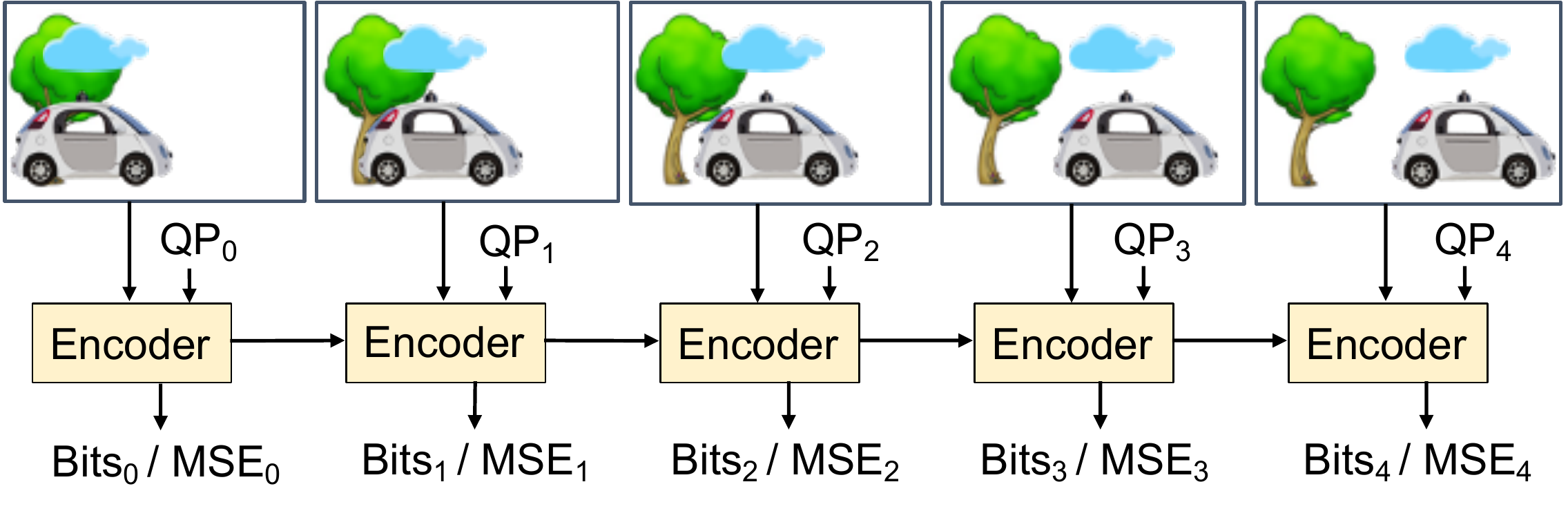}
\vspace{-1em}
\caption{Rate control via quantization parameters (QPs).}
\label{f:qp-problem}
\end{subfigure}
\hspace{0.02\textwidth}
\begin{subfigure}{0.25\textwidth}
\vspace{-0.2em}
\includegraphics[trim=0 0 0 18, clip, width=\textwidth]{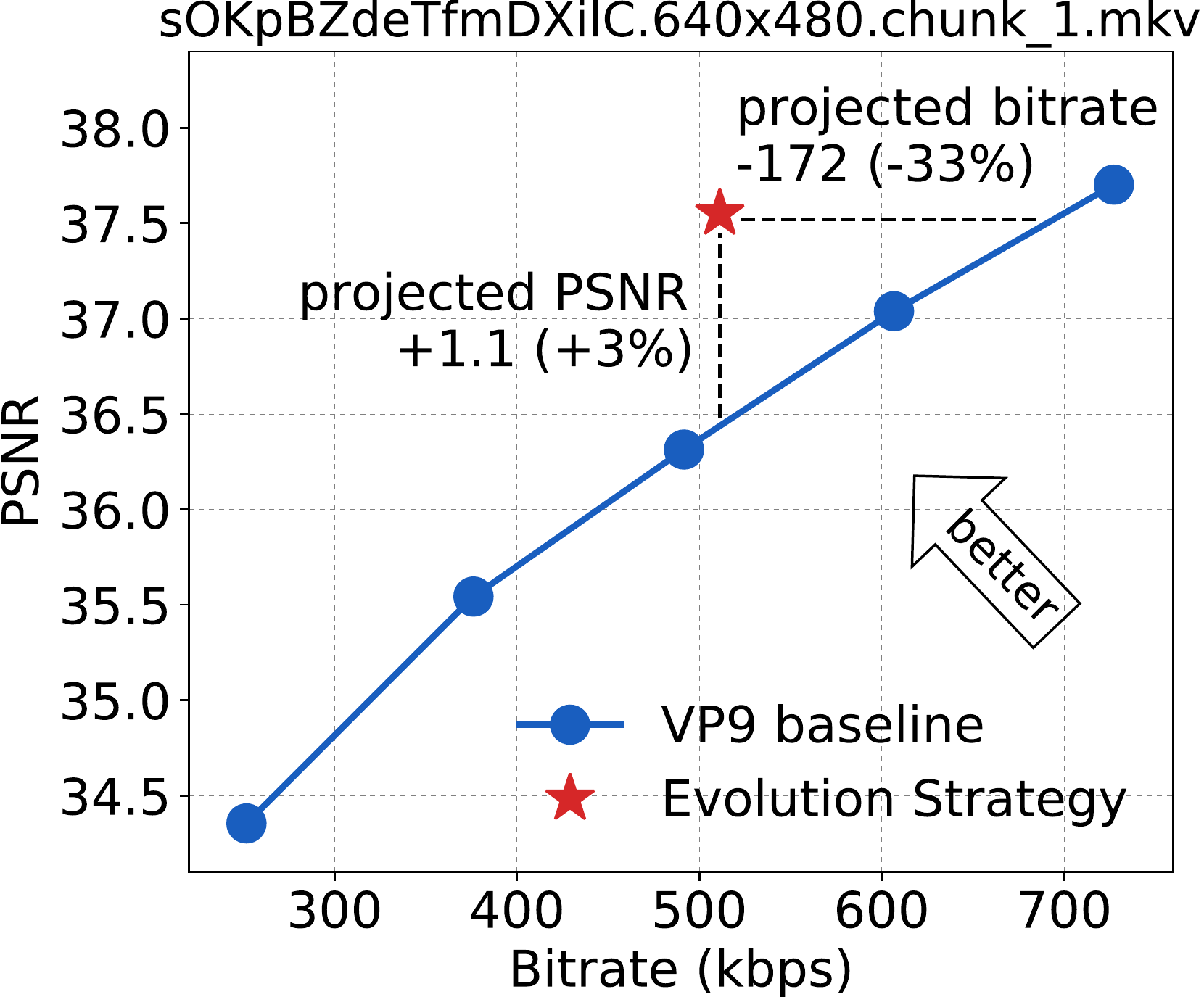}
\vspace{-1em}
\caption{Projected PSNR and bitrate.}
\label{f:iso-metric}
\end{subfigure}
\vspace{0.5em}
\caption{Illustrative example of the rate control problem and the metrics for encoding efficiency.}
\label{f:problem}
\end{figure*}

\section{Problem formulation and challenges}
\label{s:problem}

\subsection{Rate control in video encoding}

Without loss of generality, we focus on improving rate control in \libvpxx, a widely adopted open source implementation of VP9. The same methodology applies to other modern video encoders.
As illustrated in Figure~\ref{f:qp-problem}, the rate controller makes encoding decisions of video frames \emph{sequentially}.
The rate controller regulates the trade-off between rate and distortion by assigning a quantization parameter (QP) to each frame. QP is an integer in the range of $[0, 255]$. QP is monotonically mapped to a quantization step size, which is used to digitalize prediction residue for entropy coding. Smaller quantization step sizes lead to smaller quantization error but also higher bits usage. Smaller quantization error means smaller reconstruction error, measured by mean squared error (MSE).

VP9 has multiple rate control modes. In this paper, we focus on the ``two-pass, variable bitrate (VBR)'' mode~\cite{vp9-vbr} (See 
Supplementary Materials {\color{red}{A}}
for more details).
Mathematically, VBR corresponds to a constrained optimization problem (for a particular video):
\begin{equation}
\begin{aligned}
    \underset{\pi}{\text{maximize}}& \quad \text{PSNR}(\pi) \\
    \text{subject to}& \quad \text{bitrate}(\pi)  \leq  \text{bitrate}_{\text{target}}
\end{aligned},
\label{eqn:vbr}
\end{equation}
where $\pi$ is a policy that decides the QP for each video frame during the second-pass encoding.
PSNR (Peak Signal-to-Noise Ratio) measures the video quality (a.k.a., distortion) and is proportional to log-sum of mean squared error of all show frames (excluding hidden alternate reference frames). Bitrate is the sum of bits of all frames divided by the duration of the video.
Note that PSNR in \eqnref{vbr} can be replaced with other distortion metrics such as SSIM~\cite{ssim} and VMAF~\cite{vmaf}.

\ignore{
We may also extend \eqnref{vbr} by adding additional constraints such as frame buffer constraints. 

ate control would also ensure the frame size and the video buffer do not violate the video profile setting to ensure bitstream is compatible with decoder. (https://www.webmproject.org/vp9/levels/). We may want to address this in the paper by stating we can impose further constraints. 
}

\textbf{Metrics.} 
Bjontegaard delta rate (BD-rate)~\cite{bdrate} is a key metric to compare rate control policies. Given the rate-distortion (RD) (bitrate-PSNR in our case) curve of two policies, BD-rate computes the average of bitrate difference in percentage across the overlapped PSNR range, and therefore, measures the average bitrate reduction.

When we have only a single point on the RD-curve of one policy, BD-rate reduces to \emph{projected bitrate difference}, which is the difference between the bitrate of one policy and the interpolated bitrate of the other policy at the same PSNR, as illustrated in \myfigref{iso-metric}. Similarly, we define \emph{projected PSNR difference} as the difference between the PSNR of one policy and the interpolated PSNR of the other policy at the same bitrate.

\subsection{Partially Observable Markov Decision Process (POMDP) formulation}

We formulate the rate control problem \eqnref{vbr} as a POMDP.
Each step of the POMDP corresponds to the encoding of one video frame. At each step,
the \emph{action} is the QP, an integer in $[0, 255]$, for the corresponding video frame.
The \emph{state} of the MDP corresponds to \libvpxx's internal state, which is updated as new frames are encoded.
The \emph{state transition} is deterministic and is determined by how \libvpx encodes a frame.
We expose a subset of \libvpxx's internal state as the \emph{observation}, including
(1) first-pass statistics (e.g., noise energy, motion vector statistics) of all video frames,  which provides a rough estimate of the frame complexity of all frames, and helps the rate controller to plan for future frames, (2) features of the frame to be encoded (e.g., frame type and frame index) and (3) features of previously encoded frames (e.g., bits/size of the encoded frame and mean-squared error between the encoded frame and the raw frame). 
Supplementary Materials {\color{red}{B}} 
enumerates all the observations we used in this work. Note that because we do not expose all encoder states as the observation, the problem corresponds to a POMDP. However, we note that \libvpxx's default policy uses the same observations, and we find these features are sufficient to derive a good rate control policy.

To solve the constrained optimization~\eqnref{vbr}, we define the \emph{reward} to be 
\begin{equation}
R(t) = 
\begin{cases}
    0,              & \text{if } t < T \\
    \text{PSNR} - \lambda \max(0, \text{bitrate} - \text{bitrate}_{\text{target}}), & \text{if } t= T
\end{cases}
\label{eqn:reward}
\end{equation}
where $t$ is the step index, $T$ is the episode length (number of frames to encode) and 
$\lambda > 0$ is a constant that introduces a penalty term when the bitrate constraint is violated (a.k.a., bitrate overshooting).
Other reward definitions are possible. For example, we may distribute the PSNR reward over the entire episode, or introduce an overshooting penalty as soon as the bitrate constraint is violated. We leave the exploration of different reward shaping strategies for future work.

\subsection{Challenges}
\textbf{Long-horizon planning.} 
In \eqnref{reward}, PSNR is an aggregated metric over all show frames, and the bitrate is aggregated over all frames. 
Therefore, rate control corresponds to a long-horizon planning problem.
In particular, the action at the current frame affects the optimal actions for the future frames. 
For example, to exploit temporal redundancy, instead of allocating the same QP for each frame, video encoders often allocate a low QP (more bits) for the key frame such that it maintains a relatively high quality and serves as a good predictor for following frames. Consequently, the next few frames can usually be encoded with a high QP (less bits) while still achieving high quality due to the improved prediction. However, it is difficult to predict the optimal QP because of the complex dependencies of the future frame's rate-distortion tradeoff on the encoded frames.

\textbf{Episodic constraint.} Unlike common RL benchmark problems such as Atari, boardgames and robotics tasks, \eqnref{vbr} has an episodic constraint on the bitrate. The constraint is especially challenging for imitation learning, or more generally, offline (batch) reinforcement learning algorithms. Because the training does not involve interacting with the environments and evaluating policy's constraint violation, it is not obvious how the training can enforce the constraint.

\textbf{Huge search space and expensive policy evaluation.} For a video of 150 coding frames (roughly 5 seconds long), the search space of QP sequences over all frames is $256^{150}$. Like many combinatorial optimization problems, it is impractical to search for the optimal solution in a brute-force way. 
In addition, each step of the environment involves encoding a video frame, which typically takes at least a few hundred milliseconds (about 0.5 step/second). This is orders of magnitude slower than environments, such as Atari (about 6,000 steps/second) \cite{atari} and Go, and makes it harder to apply online reinforcement learning algorithms.

\textbf{Complex QP-to-bitrate, QP-to-MSE mapping.}
Intuitively, to derive a good rate control policy, the policy needs to implicitly or explicitly establish an accurate mapping from QP to the bitrate and MSE of a frame.
Existing QP-to-bitrate and QP-to-MSE models~\cite{rdmodel1, rdmodel2}  are usually over-simplified.
\libvpx works around this problem by doing a binary search to precisely find the QP for a given frame's bitrate, at the expense of additional computation and latency.

\section{Imitation learning for rate control}
\label{s:design}
\begin{figure*}[t]
\centering
\begin{subfigure}{0.46\textwidth}
\includegraphics[width=\textwidth]{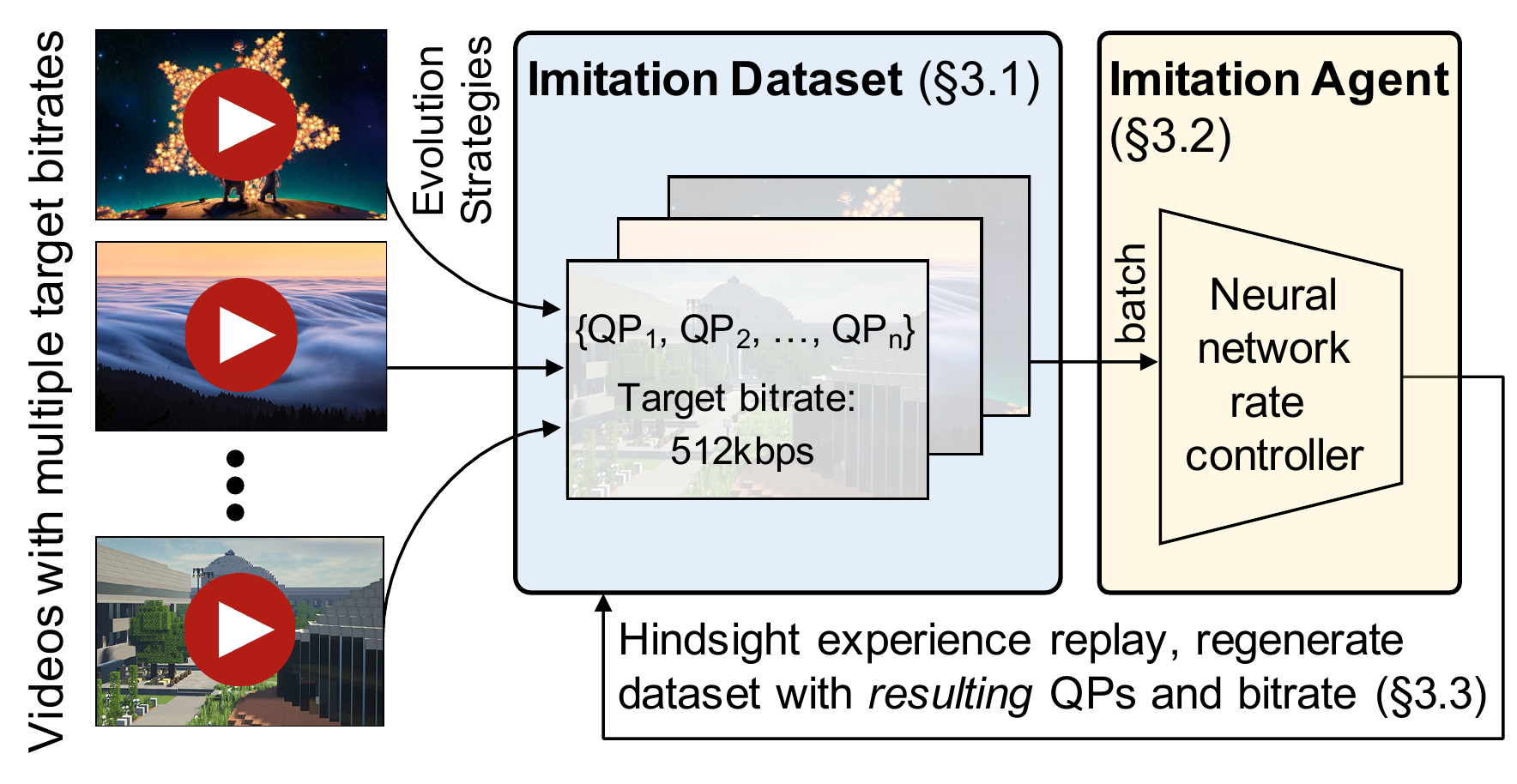}
\vspace{-1.5em}
\caption{Training.}
\label{f:overview-train}
\end{subfigure}
\hspace{0.02\textwidth}
\begin{subfigure}{0.46\textwidth}
\includegraphics[width=\textwidth]{overview-inference.pdf}
\vspace{-0.5em}
\caption{Inference.}
\vspace{-0.5em}
\label{f:overview-inference}
\end{subfigure}
\vspace{0.25em}
\caption{Overview of our method. (a) A policy network (\S\ref{s:design-model}) is trained on a teacher dataset generated by evolution strategies (\S\ref{s:design-dataset}) and refined by hindsight experience replay (\S\ref{s:design-hindsight}).
(b) At inference, low-probability QPs are truncated (\S\ref{s:design-truncation}), and the policy is augmented by feedback control to have precise bitrate control (\S\ref{s:design-control}).}
\label{f:overview}
\end{figure*}

\myfigref{overview} shows an overview of our method. 
To train a rate control policy, we first create a teacher dataset of QP sequences that perform better than the \libvpxx's rate control policy. Then we apply imitating learning to train a policy network to predict the QPs in the teacher dataset. During training, we define auxiliary losses to penalize bitrate overshooting and undershooting. We use a customized hindsight experience replay to refine the teacher dataset. During inference, we use the learned policy with a truncation trick, and we augment the policy with feedback control to further improve bitrate accuracy with little degradation in encoded video quality.

\subsection{Teacher dataset generation with evolution strategies}
\label{s:design-dataset}

Unlike problems that have human demonstration data, such as board games~\cite{alphago}, 
robotics~\cite{robot-imitation-finn}, self-driving cars~\cite{nvidia-imitation}, we do not have a readily available teacher dataset to imitate. To obtain the teacher dataset, we apply black-box optimization of \eqnref{vbr} on \emph{each} video to find a QP sequence that results in a higher reward. 
In particular, we use Evolution Strategies (ES)~\cite{es} (details in
Supplementary Materials {\color{red}{D}}) with the reward defined in \eqnref{reward}.

The teacher dataset generated from black-box optimization has a fundamental \emph{learnability} problem, for two reasons. First, multiple QP sequences can achieve similar near-optimal reward, and the optimizer may return any of these solutions. Since we apply the optimization on each video separately, the optimal solutions found for different videos may correspond to different high-level strategies, and hence makes it harder to imitate. Second, because the optimization ignores the observations and searches in the space of QP sequences (rather than the space of policies that map observations to QPs), the resulting QP sequence found by the optimization may not even be predictable from the observations.

To alleviate the learnability problem, we found it crucial to initialize ES using the QP sequence from \libvpxx's default policy. This biases ES to converge to a solution close to \libvpxx's QP sequence, which is computed from a deterministic function. Therefore, the resulting trajectories are much more coherent (compared to random initialization) and easier to imitate.

In order for the model to generalize better, we also found it important to run ES over \emph{multiple} target bitrates (10 in our experiments). This prevents overfitting to a single target bitrate. It allows the learner to see examples with a diverse distribution of \emph{remaining} bit budget, i.e., the amount of bits available to encode the rest of the frames, and therefore learn to better predict QP under different bit budgets. This is crucial to satisfy the bitrate constraint, especially for encoding the first few frames where the main information exposed to the policy is the target bitrate and the first-pass statistics, and for the last few frames where we need very precise QP prediction in order not to overshoot.

\subsection{Model architecture and training}
\label{s:design-model}

\begin{figure*}[t]
  \centering
  \includegraphics[width=0.6\textwidth]{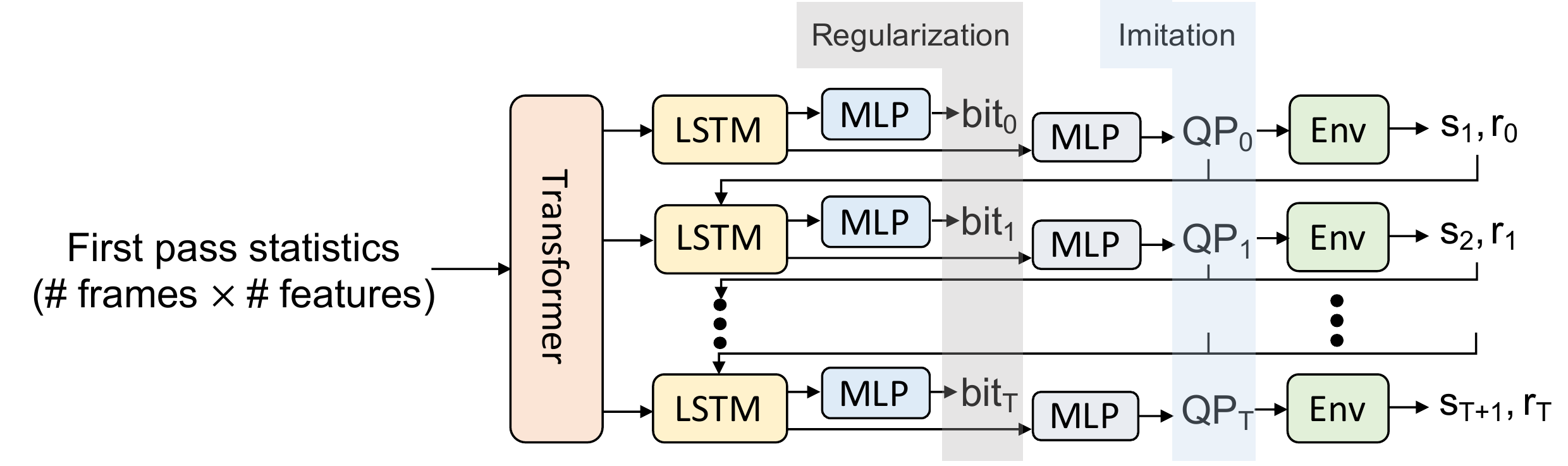}
  \caption{Policy network unrolled over $T$ video frames.}
  \label{f:imitation-model}
\end{figure*}

\textbf{Neural network architecture.}
As shown in Figure~\ref{f:imitation-model}, 
we use a transformer to process all first-pass statistics and output an embedding for every video frame.
The output of the transformer, together with the observation ($s_{t}$), reward ($r_{t-1}$), action ($\text{QP}_{t-1}$) from the previous frame, is fed to an LSTM.
The LSTM output is followed by two MLPs, one to compute logits over 256 QP values, and the other to predict the frame size $b_t$ (i.e., the number of bits of the encoded frame), which is used in auxiliary losses as regularization.
The hyper-parameters are detailed in 
Supplementary Materials {\color{red}{C}}.

\textbf{Losses.} The loss function is
\begin{equation}
    \mathcal{L} = \mathcal{L}_{\text{QP}} + \beta_1 \mathcal{L}_{\text{frame bits}}
     + \beta_2 \mathcal{L}_{\text{total frame bits}},
\label{eqn:loss}
\end{equation}
where $\mathcal{L}_{\text{QP}}$ is the cross-entropy loss between the predicted QP distribution and the imitation target,
$\mathcal{L}_{\text{frame bits}} = \sum_{t} (b_t - b_t^{\text{label}})^2$ minimizes the frame size prediction error, 
$\mathcal{L}_{\text{total frame bits}} = (\sum_{t} b_t - \text{total bit budget})^2$ minimizes the error between the total predicted bits and the total bit budget ($=\text{bitrate} \times \text{video duration}$), and $\beta_1$, $\beta_2$ are constants weighting the two loss terms.

Adding $\mathcal{L}_{\text{frame bits}}$ and $\mathcal{L}_{\text{total frame bits}}$ significantly improves model's ability to satisfy the bitrate constraint. We hypothesize that these two terms help the model learn a better mapping between QP and frame size. Since the model cannot interact with the environment during imitation learning, these two losses are the only way for the model to understand its behavior related to the bitrate constraint. 
In particular, we found $\mathcal{L}_{\text{total frame bits}}$ to be especially important. Without this loss term, the bits prediction error often have the same sign for all frames in a video, leading to systematic under-prediction/over-prediction and therefore significant bitrate undershooting and overshooting.

\subsection{Dataset refinement with hindsight experience replay (HER)} 
\label{s:design-hindsight}

Training the policy network using the loss \eqnref{loss} on the ES teacher dataset in \S\ref{s:design-dataset} achieves almost 10\% reduction in projected bitrate. However, the policy satisfies the target bitrate constraint for only about 20\% of the test videos.

We hypothesize this is partially due to the learnability of the teacher dataset. As mentioned in \S\ref{s:design-dataset}, the optimal QP sequences in the teacher dataset are obtained by running ES on each video individually. 
However, for each video, there exist multiple QP sequences that lead to similar encoding metrics, and ES can return any one of these solutions. As a result, the QP sequences in the dataset can correspond to different underlying strategies.

To make the training data more coherent, we use the learned policy to generate a new teacher dataset. 
However, because most trajectories violate the bitrate constraint in \eqnref{vbr}, we cannot train on this dataset directly.
Instead, for each trajectory, we modify the target bitrate (which is an input to the policy) to the actual bitrate at the end of of the episode.
This is similar to hindsight experience replay~\cite{her} which modifies the goal of the agent to be the goal actually achieved in the episode. In our problem, the goal is the target bitrate, which is defined in the constraint in \eqnref{vbr}.
Figure~\ref{f:ablations} shows that training on the HER-generated dataset improves both encoding efficiency and bitrate accuracy.

\subsection{Truncation trick to improve model stability}
\label{s:design-truncation}

During inference, instead of sampling from the predicted QP distribution, we truncate the logits and keep only top 15 logits and their QP values. Then we sample a QP from this truncated distribution.

This design choice is based on two empirical observations.
First, the top 15 QPs predicted by a trained policy network cover more than $80\%$ of the QP labels.
Second, occasionally the learned policy samples a (catastrophic) QP that is far away from the optimal QP,  because the model has a large error/uncertainty at these frames. This causes the video encoder assigning too many or too few bits to a particular frame. Not only does this lead to low encoding efficiency for the frame, it also drives the rest of the episode into a region of the state space that is not well covered in the training data. With the truncation trick, we effectively avoid these obviously bad actions. Empierically, we observe much more stable and better metrics.

\subsection{Policy augmented with feedback control}
\label{s:design-control}

\begin{figure*}[t]
\centering
\begin{subfigure}{0.38\textwidth}
\includegraphics[width=\textwidth]{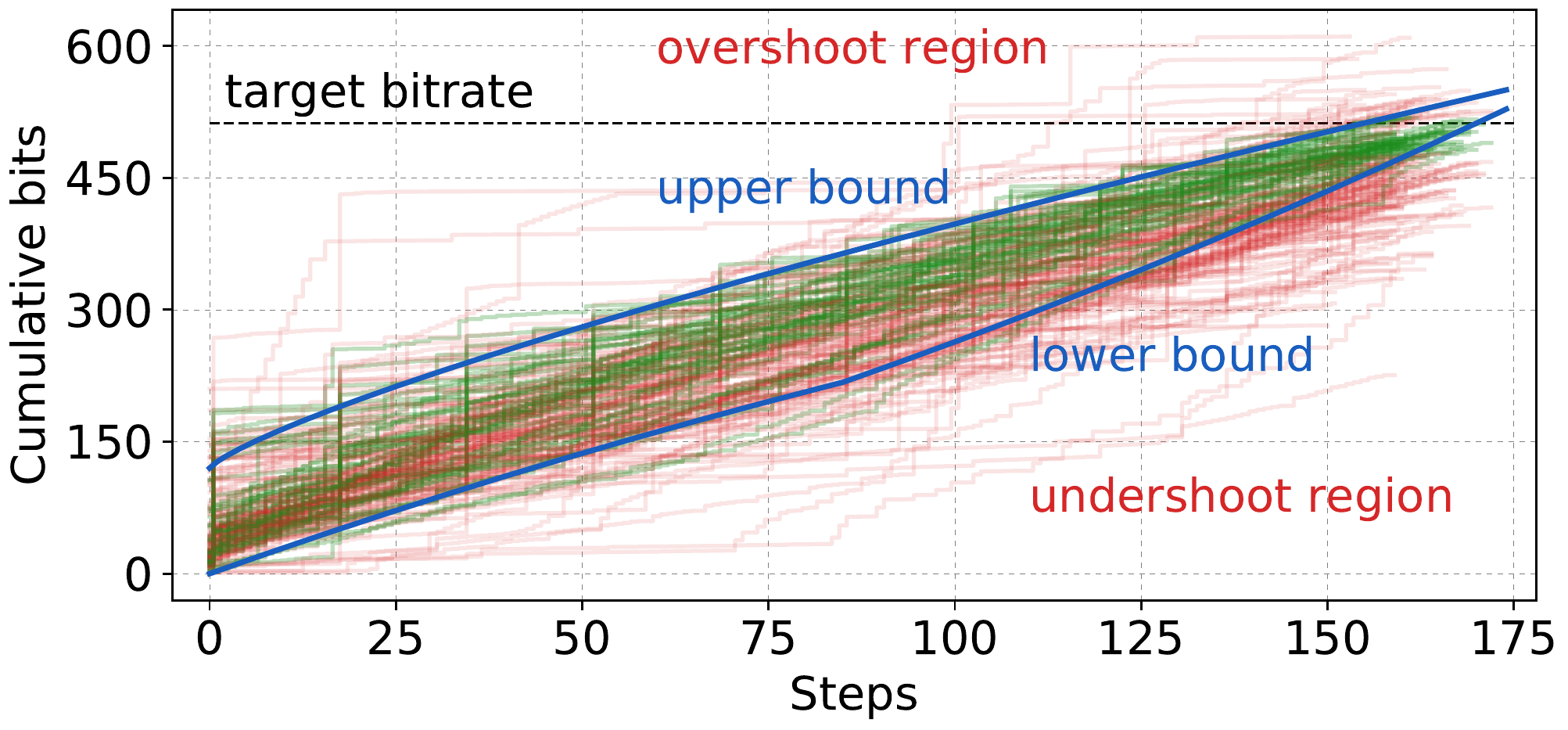}
\vspace{-1.5em}
\caption{Bitrate trajectories.}
\label{f:control-boundary}
\end{subfigure}
\hspace{0.02\textwidth}
\begin{subfigure}{0.48\textwidth}
\includegraphics[width=\textwidth]{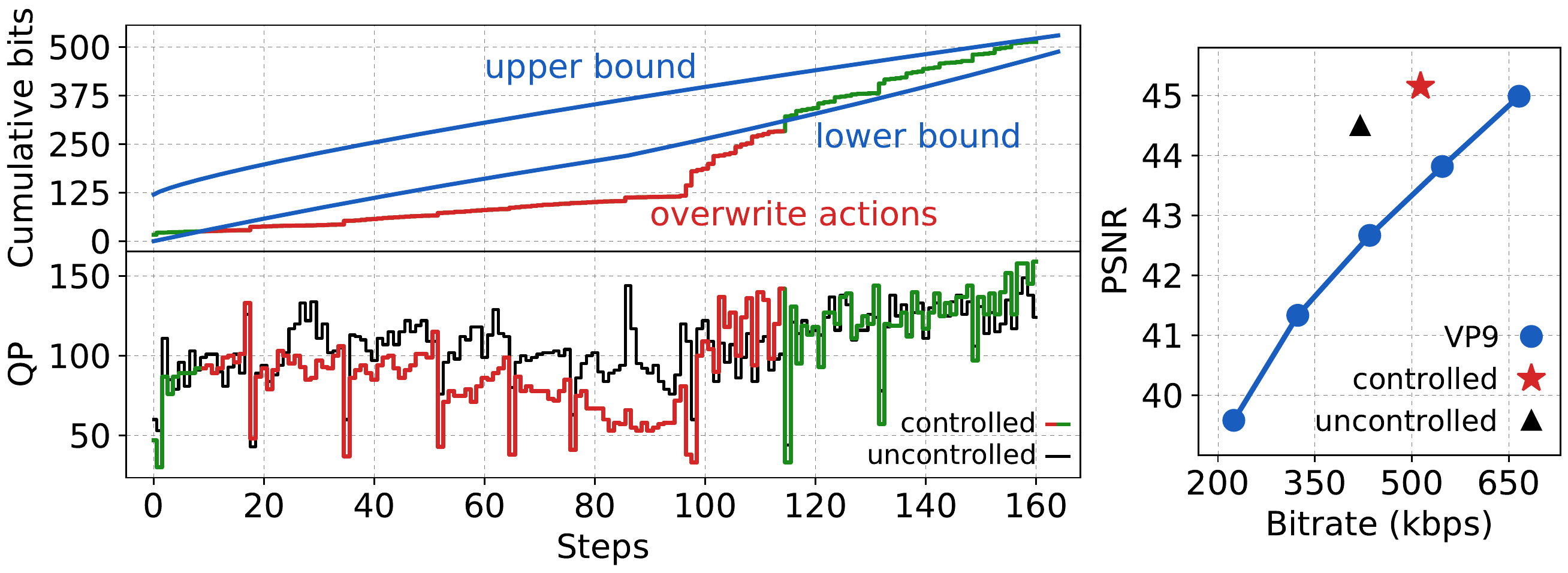}
\vspace{-1.5em}
\caption{Feedback control and its effect.}
\label{f:control-example}
\end{subfigure}
\vspace{0.25em}
\caption{Illustration of QP feedback control. (a) Upper/lower bounds of cumulative bits, overlaid with trajectories of 500 episodes. (b) An episode where the feedback is triggered between step 9 and step 115.}
\vspace{-0.25em}
\label{f:feedback-control}
\end{figure*}

With the auxiliary losses, HER and the truncation trick, the policy satisfies the bitrate constraint for more than 50\% videos. However, we still observe episodes with significant bitrate undershoot or overshoot.
We believe this is because during inference, there are always inputs for which the model prediction is not good, and when the model predicts a bad action, the state enters a region where the model makes systematic errors and it is hard for the model to self-correct.

To address this problem, we augment the learned policy with a classical feedback control mechanism to steer the policy behavior during inference. As shown in \myfigref{overview-inference},
the feedback control has two components: (1) pre-defined lower/upper bounds of bits consumption over encoding steps and (2) QP offset proportional to the deviation from the lower/upper bounds. When the bits consumption is out of the lower/upper bounds, we offset the predicted QP (from top-15 sampling) by an amount proportional to the difference to the bounds. This forces the policy to stay within the lower/upper bounds, and eventually hit the bitrate target much more precisely, even if the model itself is inaccurate.

Specifically, let $B^{\text{lower}}_t$ and $B^{\text{upper}}_t$ be the lower and upper bounds of bits consumption at time $t$, and $B_t$ be the actual bits consumption. First, like the truncation trick, we select top 40 QP values, $\text{QP}_1, \text{QP}_2, \cdots, \text{QP}_{40}$ as candidate QP values. Suppose the QP from top-15 sampling is $\text{QP}_i$, and the adjusted QP is $\text{QP}_j$, we have
\begin{equation}
j = 
\begin{cases}
    \min(1, i - \alpha (B^{\text{lower}}_t - B_t),              & \text{if } B_t < B^{\text{lower}}_t \\
    i, & \text{if } B^{\text{lower}}_t \le B_t \le B^{\text{upper}}_t \\
    \max(40, i + \alpha (B_t - B^{\text{upper}}_t),              & B_t > B^{\text{upper}}_t
    \end{cases},
\label{eqn:feedback}
\end{equation}
where $\alpha$ is a constant adjusting the strength of the feedback control. We found that limiting the control in the top-40 QPs is important for the policy to still achieve better encoding efficiency.
We compute $B^{\text{lower}}_t$ and $B^{\text{upper}}_t$ so that (1) most trajectories in the training data fall within the bounds, and (2) the bound is tighter towards the end of the episode. This is illustrated in Figure~\ref{f:control-boundary} and detailed in Supplementary Materials {\color{red}{C.3}}.
Intuitively, the feedback control makes sure the inputs to the model stay close to those in the training data where the model have good prediction accuracy.

Figure~\ref{f:control-example} shows a working example of feedback control.
At step 9, $B_t$ is under $B_t^{\text{lower}}$ and the feedback control is triggered and lowered the predicted QP. At step 115, $B_t$ goes back to the range $[B_t^{\text{lower}}, B_t^{\text{upper}}]$ and the feedback control is disabled. 
In this example, the augmented policy reaches a bitrate within 2\% of the target bitrate (compared to 24\% of the policy without feedback), and achieves the same improvement in encoding efficiency.

The above feedback control can be thought of as a \emph{policy improvement operator} which improves the policy by reducing constraint violations. 
While it is designed for the rate control problem, it is generic and can handle other problems where (1) the episodic constraint is additive and (2) the action is monotonic w.r.t. the quantity in the constraint.
It may also be used in a policy-evaluation-policy-improvement loop, to allow training focus on policies that respect the constraints.

\section{Experimental results}
\label{s:eval}

\textbf{Experiment setup.} We use \libvpxx, an open source VP9 codec library, in our experiments. We modified \libvpx to allow a machine learning model to take over QP assignment in the rate control module. We use the two-pass VBR mode with speed 0, which achieves \libvpxx's best encoding efficiency.

We train and evaluate on a randomly sampled set of videos, spanning diverse content types (games, natural scenes, etc) and spatial/temporal complexities (different textures, fast/slow motions). For simplicity, we constrained our study for videos of 640$\times$480 resolution, 100-150 frames long and have a baseline PSNR in $[25, 45]$ (effectively excluding very easy and very hard to encode videos). We train on 1600 videos, each encoded at 10 random target bitrates in $[256\text{kbps}, 768\text{kbps}]$. We evaluate the policy at target bitrate 512kbps, on a held-out set of 200 videos, and compare it to \libvpxx's rate control policy (baseline). We evaluate two main metrics: (1) reduction in projected bitrate (approximation to BD-rate) and (2) deviation of bitrate from the target bitrate (measuring constraint violations).


\begin{figure*}[tbp]
\centering
\includegraphics[width=0.85\textwidth]{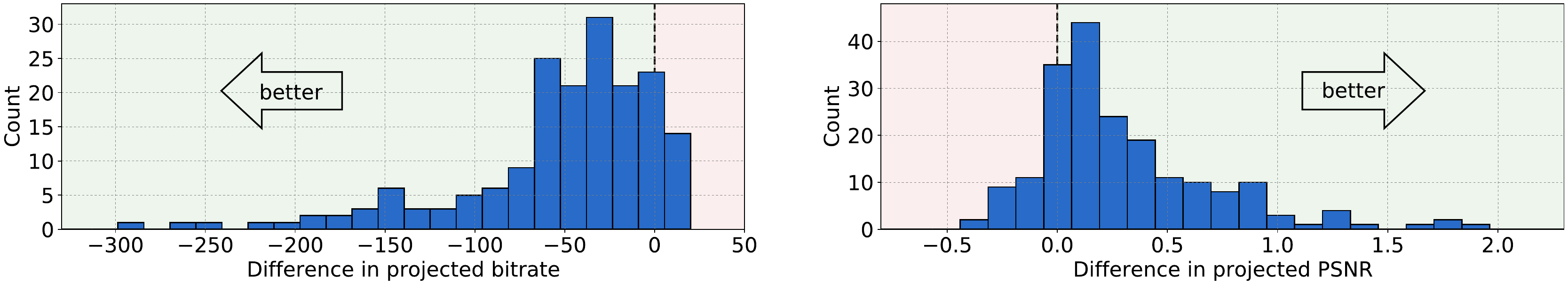}
\vspace{-0.5em}
\caption{Histogram of the difference in projected bitrate/PSNR betweeen our policy and \libvpxx's policy.}
\label{f:overall-projected-metrics}
\end{figure*}

\begin{figure*}[tbp]
\centering
\includegraphics[width=1\textwidth]{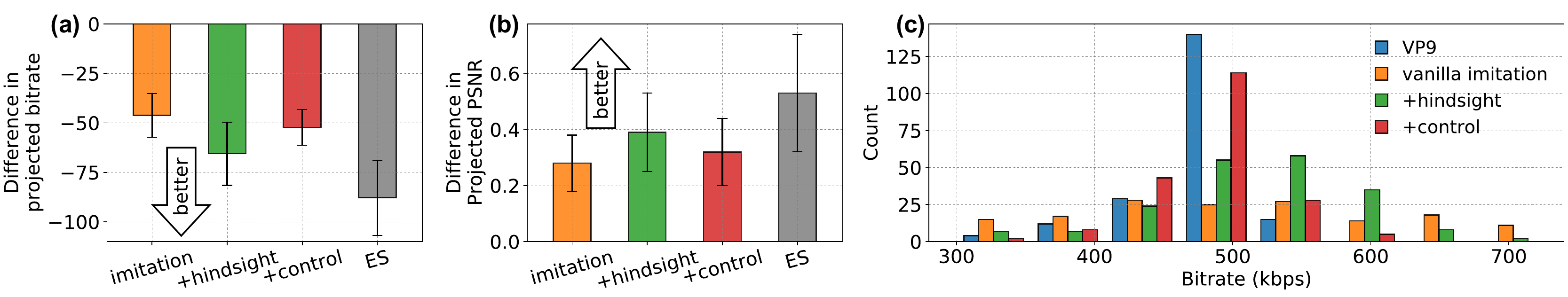}
\vspace{-1.75em}
\caption{Ablation of HER and feedback control. Results of ES and \libvpxx's policy are included as baselines.}
\label{f:ablations}
\end{figure*}

\textbf{Summary of results.}
Applying ES on individual videos, we observe an average 13\% improvement in BD-rate. This is a rough estimate of how much we can improve \libvpxx's QP rate control policy. 
Figure~\ref{f:ablations}{\color{red}a} and {\color{red}6b} show the improvement in projected bitrate/PSNR of ES.
With $\lambda = 0.02$ in \eqnref{reward}, the bitrate is always close to the target bitrate.

Figure~\ref{f:overall-projected-metrics} shows the histogram of projected bitrate/PSNR differences between the imitation learning policy and the baseline policy. The median reduction in projected bitrate is 8.5\% (43.5kbps in absolute value) -- for about 80\% of the videos, we observe up to 10\% reduction. Similarly, the median improvement in projected PSNR is 0.22. For most videos, the learned policy saves up to 10\% of the encoded video size, and achieves the same PSNR, leading to savings in both bandwidth and storage.

Figure~\ref{f:ablations}{\color{red}c} shows the bitrate distribution of encoded videos, using \libvpxx's policy and different variants of the learned policy. The percentage of videos whose bitrate is smaller than 520kbps is 86\% (learned) and 99\% (baseline), respectively. The percentage of videos whose bitrate is in $[480, 520]$kbps is 64\% (learned) and 70.5\% (baseline). The bitrate accuracy is very close to that of the baseline (although slightly worse), and the accuracy is good enough for most applications.

The latency of our model is 15ms on a single skylake CPU. For comparison, the latency for encoding a frame is more than 300ms. Therefore, the overhead in computation/latency is minimal and is outweighted by the BD-rate improvement.

\textbf{Ablation study.} We study how HER (\S\ref{s:design-hindsight}) and feedback control (\S\ref{s:design-control}) affect the final model performance. Figure~\ref{f:ablations} shows the contribution of each factor for both projected metrics and bitrate distribution. Refining the dataset with HER makes the target easier to mimic, and improves the projected metrics over the policy directly on the teacher dataset from ES. The feedback control leads to a slight degradation in projected bitrate/PSNR, because the control mechanism overrides some good actions predicted by the model. However, it significantly improves the accuracy of the final bitrate and minimizes bitrate overshooting. 

\begin{figure*}[tbp]
\centering
\includegraphics[width=\textwidth]{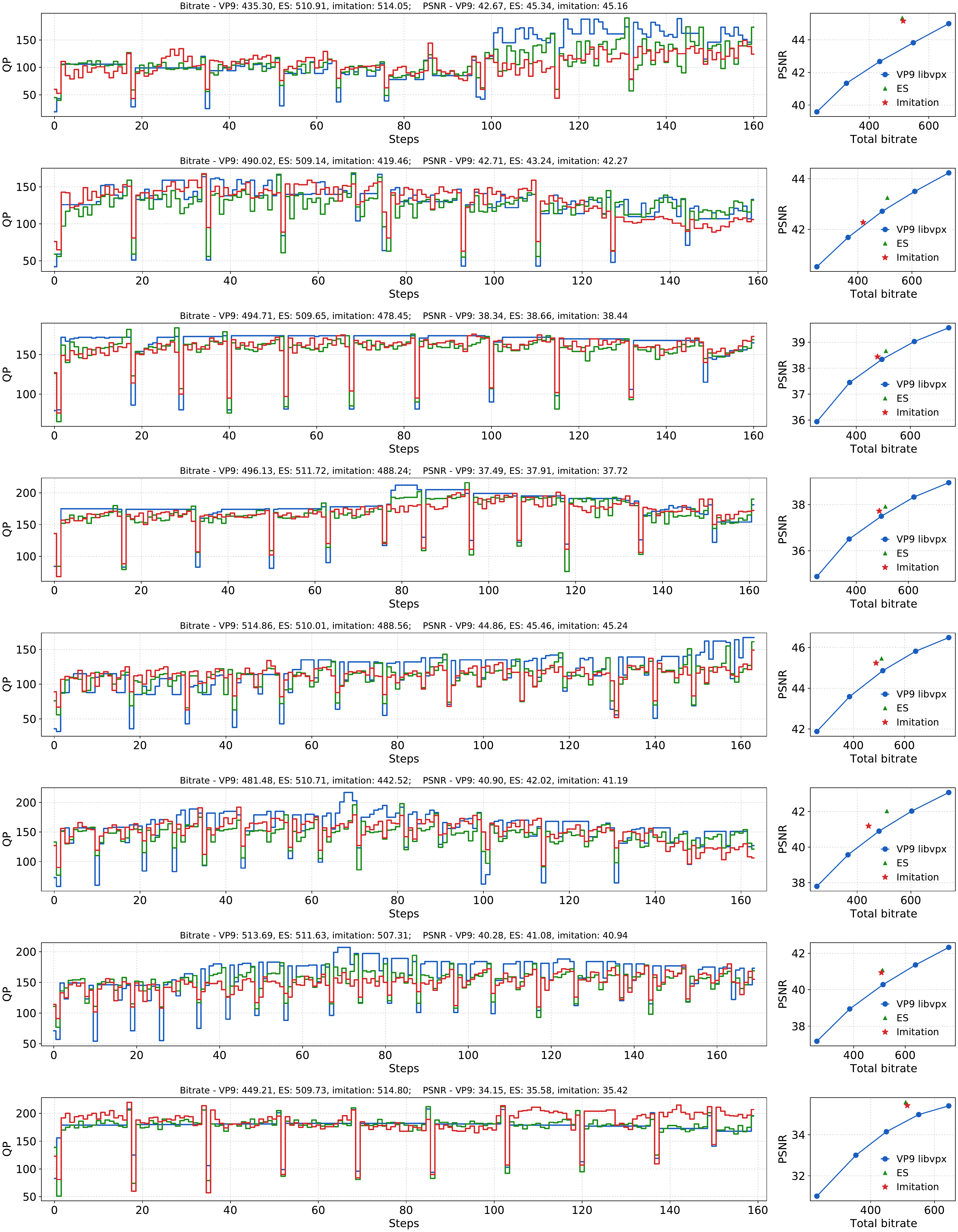}
\caption{Samples of QP sequences comparing our learned policy, \libvpxx's policy and ES.}
\label{f:qp-sequence-example}
\end{figure*}

\textbf{Analysis of encoding trajectories.} 
Figure~\ref{f:qp-sequence-example} visualizes a few QP trajectories by our learned policy, ES, and \libvpxx's policy.
We observe that \libvpx's policy tends to aggressively allocate lots of bits (low QP) to the first key frame and the hidden alternate reference frames, because they are heavily used as reference frames. However, ES often discovers a QP sequence that does not assign as much bits to these frames, and imitation learning is able to learn that strategy. The learned policy distributes the excessive bits saved from key/reference frames to improve quality of other frames, and hence the overall quality.

\section{Related Work}
\label{s:background}
\textbf{Rate control in video codecs.} 

Traditionally, rate control algorithms are based on empirical models using relevant features, such as mean absolute differences (MAD) and sum of absolute transformed differences (SATD), from past encoded frames. Common empirical models include quadratic models used in MPEG-4's VM8~\cite{derek2} and H.264/AVC~\cite{derek3,derek5,derek6}, $\rho$-domain models that map between rate and the proportion of non-zero quantized coefficients~\cite{derek7}, and dynamic programming based models that exploit interframe dependency~\cite{derek8}.  In modern codecs, such as HEVC and VVC, \cite{rdmodel2} introduced $\lambda$-domain model to control both rate and mode selections through rate-distortion optimization instead of relying on QP decisions. Despite recent advances in rate control models, the empirical rate control models could not effectively adapt to different dynamic video sequences. 

\textbf{ML for rate control.}
Machine learning enables more complex and nonlinear models to be formulated. \cite{derek10} uses a radial basis function network to predict QP offset and ensure quality consistency. \cite{derek11} utilizes a game theory method to allocate CTU-level bit allocation and optimize for SSIM in HEVC. Reinforcement learning approaches in rate control is also proposed to HEVC recently 
in~\cite{derek12,derek13,rl-rate-control-2020-ieee}.

\textbf{Neural video compression.}
Neural video compression is an active area of research. Neural networks have been used as autoencoders with quantized latent variables~\cite{rdae}, frame interpolation models \cite{interp1, interp2} and optical flow estimation and motion compensation models \cite{cvpr-video-encode,rippel2019learned,scalespacee2e}.
These methods often involves end-to-end differentiable neural networks and/or require changes in the decoder. In contrast, our method in this paper  enhances \libvpxx's rate control, and hence is much less intrusive and is readily deployable.

\section{Conclusion}
\label{s:conclusion}

We have demonstrated that with imitation learning, a neural network rate control policy can be learned to achieve better encoding efficiency compared to the rate control policy in  \libvpxx, a widely adopted VP9 codec library. Our approach does not require any change in the decoder, and hence is readily deployable.
While we focused on the rate control problem, we believe techniques developed in this paper are general and are applicable to other imitation learning problems with episodic constraints.
We also believe the same method is applicable to other state-of-the-art video encoders and can generally improve encoding efficiency.

\section{Acknowledgment}

We would like to thank Daniel Visentin, Chris Fougner, Chip Schaff, Alex Ramirez, Anton Zhernov, Amol Mandhane, Maribeth Rauh for fruitful discussions and their contributions to the project. We also thank Luis Cobo, Harish Chandran, Dilan Gorur, Balu Adsumilli, Nishant Patil, Yaowu Xu for the support throughout the project.

\newpage 
{\small
\bibliographystyle{ieee_fullname}
\bibliography{paper}
}

\appendix
\newpage
\noindent
\textbf{\Large Appendices}
\section{VP9's two-pass VBR mode}
\label{a:vbr}

In this section, we briefly describe how VP9's two-pass VBR mode works.

In the first pass encoding, the encoder computes statistics for every frame in the video, by dividing the frame into non-overlapping $16\times 16$ blocks followed by per-block intra- and inter-frame prediction and calculation of prediction residual error.
The statistics contain information such as average motion prediction error, average intra-frame prediction error, average motion vector magnitude, percentage of zero motion blocks, noise energy, etc.

In the second pass encoding, the encoder uses the first-pass stats to decide keyframe locations and insert hidden alternate reference frames.
With those decisions made, the encoder starts to encode video frames sequentially.
The rate controller regulates the trade-off between rate and distortion by specifying a quantization parameter (QP) to each frame.
The quantization parameter is an integer in range [0, 255] that can be monotonically mapped to a quantization step size which is used for digitalize prediction residue for entropy coding. The smaller quantization step size leads to smaller quantization error but also higher bits usage for the frame.
Smaller quantization error means smaller reconstruction error which is measured by mean squared error (MSE) and is used to compute PSNR to measure the encoded video quality.

\section{MDP observations}
\label{a:state-details}

We expose the following features as the MDP \emph{observation} that is fed to the policy network.

\textbf{Static features.} Static features include video metadata (resolution, number of frames, duration, frame rate) and encoding settings (target bitrate, encode speed).

\textbf{First-pass features.} We extract 25 features for each frame from first-pass encoding in \libvpxx. As shown in Table~\ref{table_spf}, the features include different coding errors, frame noise energy, percentage of blocks that use inter/intra modes, and statistics of motion vectors.  These features measure the spatial and temporal complexity of the frames, and are critical for the policy to plan for the future.
The first pass statistics of all the frames are available before the episode (second-pass encoding) begins, and can be viewed as contextual features that vary from video to video.

\begin{table*}[h!]
\begin{center}
\begin{tabular}{ l p{10cm} } 
\hline
Feature  & Description \\
\hline
frame\_index  & Frame number in display order \\
frame\_weight & Frame weight used to differentiate easy and hard frames \\
intra\_error & Intra-frame prediction error \\
coded\_error & Best of intra-frame prediction error and inter-frame prediction error using last frame as reference \\
sr\_coded\_error &
Best of intra-frame prediction error and inter-frame prediction error using Golden frame as reference \\
frame\_noise\_energy &
Noise energy estimation \\
pcnt\_inter & 
Percentage of blocks with inter pred error $<$ intra pred error \\
pcnt\_motion & 
Percentage of blocks using inter-frame prediction and non-zero motion vectors \\
pcnt\_second\_ref &
Percentage of blocks where golden frame was better than last frame or intra-frame \\
pcnt\_neutral &
Percentage of blocks where intra and inter prediction errors were very close \\
pcnt\_intra\_low &
Coded with intra-frame prediction but of low variance \\
pcnt\_intra\_high &
Coded with intra-frame prediction but of high variance \\
intra\_skip\_pct &
Percentage of blocks that have almost no intra error residual \\
intra\_smooth\_pct &
Percentage of blocks that are smooth \\
inactive\_zone\_rows &
Image mask rows top and bottom \\
inactive\_zone\_cols &
Image mask columns at left and right edges \\
MVr &
Average of row motion vectors \\
mvr\_abs &
Mean of absolute value of row motion vectors \\
MVc &
Average of column motion vectors \\
mvc\_abs & 
Mean of absolute value of column motion vectors \\
MVrv &
Variance of row motion vectors \\
Mvcv &
Variance of column motion vectors \\
mv\_in\_out\_count &
Fraction of row and column motion vectors that point inwards or outwards \\
duration &
Duration of the frame / collection of frames \\
frame\_count & Frame count \\
\hline
\end{tabular}
\end{center}
\caption{First-pass features.}
\label{table_spf}
\end{table*}

\textbf{Second-pass features of the to-be-encoded frame.} Before encode frame $t$, we extract its frame type (e.g., key frame, alternate reference frame, inter frame), and frame index. 

\textbf{Second-pass features of encoded frames.} After encoding frame $t$, we extract features from the encoded frame as well as statistics used during encoding.
The most important features are the frame size (number of bits used to encode the frame), and mean squared error between the encoded frame and the original frame. 
Other features include various counters, such as counts of block sizes, transform sizes, inter/intra modes, skipped blocks, motion vector types, etc.

\section{Model details and hyperparameters}
\label{a:hyper-params}

\subsection{Inputs to the policy network}
The inputs to the policy network include observations as well as the last action (QP) and metadata of the video/encoding setting. The reward is not used directly, but the observations contain sufficient statistics to compute the reward.

We normalize most float features to have zero-mean and unit-variance. We apply $\log(1+\cdot)$ to count features.
We build 16-dimensional embeddings for each QP value and 16-dimensional embeddings for the frame type.

We define absolute cumulative bits and relative cumulative bits (w.r.t. bit budget defined by the target bitrate), as they explicitly indicate how many bit budget is left and is useful for the policy to plan to avoid bitrate violation.

Note that while we have access to the original and encoded video frame image, we do not use them as the input to the model in this work. We could potentially use any neural network to process the video frames to further improve the policy.

\subsection{Policy network and training hyper-parameters}

In the policy network, we use a single-layer transformer with 16 heads, key/value/query sizes 16, hidden output size 128, dropout rate 0.1, with layer normalization at the input and relative position encoding.
The LSTM has 128 hidden units. 
The MLPs after the LSTM has two layers with number of hidden units 32 and 16, respectively. 

The regularzation coefficients in Equation~({\color{red}{3}}) are $\beta_1=2$ and $\beta_2 =2$.

The model is trained with teacher-forcing, i.e., we use the ground-truth actions/observations of steps $1:t-1$ to predict the action at step $t$.

\subsection{Fitting lower and upper bounds for feedback control}
\label{a:fitting-bounds}

We compute $B^{\text{lower}}_t$ and $B^{\text{upper}}_t$ in Equation~({\color{red}{4}}) so that (1) most trajectories in the training data fall within the bounds, and (2) the bounds are loose in the middle of the episode (to account for variations of videos), and tight towards the end of the episode (to precisely reach the target bitrate). 
To do that, we fit two parameterized logarithmic functions as the boundary of the
envelope (in the form of $a_1\log(a_2x + a_3) + a_4x + a_5$). However, one can use any parameterized function (e.g., a piecewise linear function) to fit the upper/lower bounds, and it would also work fine.

Figure~{\color{red}{4}} visualizes 500 cumulative bits (divided by the duration of the video) trajectories, corresponding to encoding episodes of 500 videos at target bitrate 512kbps.
We color the trajectories green when they hit $\pm 5\%$ of the target
bitrate and color them red otherwise.
Staying within the upper/lower bounds is a strong indicator for a typical encoding episode that does not overshoot/undershoot. Around the middle of the episodes, $72\%$ of the 
trajectories that stays between the bounds eventually hit (within $\pm 5\%$ of) the
target bitrate. Similarly, $89\%$ of the trajectories
that falls outside the boundary eventually miss the target bitrate.

\section{Black-box optimization using evolution strategies}
\label{a:es}

To generate the teacher dataset, 
We use evolution strategies to generate the teacher dataset and to evaluate the room for improvement of \libvpxx's rate control policy.

We use the standard ES update rule
\begin{equation}
\theta_{t+1} \leftarrow \theta_t + \alpha \frac{1}{n\sigma}F(\theta_t + \sigma \epsilon_i) \epsilon_i
\end{equation}
where $\alpha$ is the learning rate, $n$ is the batch size, $F(\cdot)$ is the reward function, $\theta$ is the parameters and $\sigma$ defines the standard deviation of the Gaussian noise $\epsilon_i$ that defines the perturbation of the parameters.

In our setup, we do not learn the parameters of a neural network. Instead, we look for a QP sequence that achieves a better encoding efficiency. Therefore, $\theta = [\text{QP}_0, \text{QP}_1, \cdots, \text{QP}_{T-1}]$ for an encoding episode of $T$ frames. Because $\text{QP}_i$ is an integer value in the range of $[0, 255]$, we apply rounding after the Gaussian noise is added at every ES step.

We use Equation~({\color{red}{2}}) as the reward function and use $\lambda=0.02$ to penalize bitrate overshooting.
We use $\sigma=4$.
We use an initial learning rate of 16.0 with exponential decay schedule at 0.5 decay rate per 100 steps.
We use batch size 16 and train up to 300 steps. For most videos, ES reaches a near-optimal reward after 100 steps.

Most importantly, we initialize the ES with the QP sequence of \libvpxx's policy.
We found that with random initialization, ES tends to find a QP sequence that is flat. While these sequences are equally good, or even better, in terms of encoding efficiency, they are difficult to imitate. We believe this is because ES does not map observations to QP, and therefore may find a QP sequence that is good in performance but is not predictable from the observations.

\end{document}